\documentclass{article}
\usepackage[final]{neurips_2020}

\usepackage[utf8]{inputenc} 
\usepackage[T1]{fontenc}    
\usepackage{hyperref}       
\usepackage{url}            
\usepackage{booktabs}       
\usepackage{amsfonts}       
\usepackage{nicefrac}       
\usepackage{microtype}      
\usepackage{amsmath}
\usepackage{amssymb}
\usepackage{wrapfig}
\usepackage{floatrow}
\usepackage{graphicx}
\usepackage{float}
\usepackage{xcolor}
\usepackage{enumitem}

\makeatletter
\newcommand{\printfnsymbol}[1]{%
  \textsuperscript{\@fnsymbol{#1}}%
}
\makeatother

\title{Structural Forecasting for Tropical Cyclone Intensity Prediction: Providing Insight with Deep Learning}

\author{%
  Trey McNeely$^1$ $\quad\quad$ Niccol\`o Dalmasso$^1$ $\quad\quad$  Kimberly M. Wood$^{\, 2}$ $\quad\quad$  Ann B. Lee$^{\, 1}$ \\
  \\
  $^{1}$Department of Statistics \& Data Science, Carnegie Mellon University \\
  $^{2}$Department of Geosciences, Mississippi State University \\
  \texttt{imcneely@stat.cmu.edu} 
}

\begin{document}

\maketitle
\vspace{-0.5cm}
\begin{abstract}
Tropical cyclone (TC) intensity forecasts are ultimately issued by human forecasters. The human in-the-loop pipeline requires that any forecasting guidance must be easily digestible by TC experts if it is to be adopted at operational centers like the National Hurricane Center. Our proposed framework leverages deep learning to provide forecasters with something neither end-to-end prediction models nor traditional intensity guidance does: a powerful tool for monitoring high-dimensional time series of key physically relevant predictors and the means to understand how the predictors relate to one another and to short-term intensity changes.
\end{abstract}
\vspace{-.25cm}

\section{Introduction}
Tropical cyclones (TCs) are powerful, highly organized storm systems that help transfer energy from the upper levels of the world’s oceans to the atmosphere. Increasing coastal populations, concurrent with rising ocean temperatures due to climate change, increase the dangers (such as storm surge and damaging winds) posed by TCs. Thus, accurate prediction of TC trajectories and intensities becomes an ever more critical component of disaster preparation and response. Track forecasting has made great strides since the 1990s, but intensity forecasting, especially in the short-term 12-h and 24-h windows, have seen less pronounced improvements \cite{DeMaria2014,cangialosi2020}. Cases of rapid intensity change (30-knot intensity changes in 24 hours), in particular, have proved difficult to predict \cite{Kaplan2010,kaplan2015rii,wood2015definition}. 

TC prediction and science rely in part on the relationships between the external environment, internal structure, and behavior of these storms. For example, vertical wind shear (an external factor) and TC convective patterns (an internal factor) are both known to relate to TC intensities \cite{Dvorak1975,kaplan2015rii,hu2020short}. Forecasters can access a wide array of real-time TC observations of both types, but new public forecasts must be issued every six hours by expert human forecasters. This human-in-the-loop pipeline requires that any forecasting guidance must be easily understood by TC experts. If the output of a method cannot be digested by scientists in a handful of minutes, it will not be adopted by stakeholders at operational centers such as the National Hurricane Center (NHC).

The NHC and other operational forecast centers use both physics-based models (dynamical models) and data-driven models (statistical models) \cite{cangialosi2020}. The Statistical Hurricane Intensity Prediction Scheme (SHIPS \cite{DeMaria1999}) is a particularly successful statistical-dynamical model which largely relates area averages of environmental fields (such as vertical wind shear) and cloud-top temperatures from infrared (IR) imagery to future TC intensity change. This IR imagery is available at high spatial and temporal resolutions from satellites like the Geostationary Operational Environmental Satellites (GOES\footnote{https://www.star.nesdis.noaa.gov/goes/} \cite{schmit2017closer,Knapp2018}). Structure in IR imagery has long been known to relate to TC intensity (see the Dvorak technique \cite{Dvorak1975,olander2019advanced}), but the area-averaged values commonly used in statistical models discard critical spatial information. The evolution of spatial structure in such fields correlates with intensity change, but unlocking this rich source of information in IR imagery without sacrificing interpretability remains an open problem.

We propose a framework to incorporate evolving spatial information into intensity forecasts by utilizing deep learning (DL) in the high-dimensional time series setting while remaining cognizant of the needs of the end users: forecasters and scientists. While our end goal is the prediction of short-term intensity changes, our framework will offer something neither end-to-end models (black-box prediction of intensity directly from IR imagery) nor existing operational forecasts can: {\bf structural summaries} or interpretable quantification of the convective structure of TCs in the form of one-dimensional functions, together with {\bf structural forecasts} of the dynamic evolution of these summaries. This new setup will provide forecasters with a powerful tool for monitoring high-dimensional time series of key physically relevant predictors (like the eye-eyewall structure and symmetry of deep convection relative to the TC center) with the means to understanding how the predictors relate to one another and to TC intensity change.

\begin{figure}
\floatbox[{\capbeside\thisfloatsetup{capbesideposition={left,top},capbesidewidth=.24\textwidth}}]{figure}[\FBwidth]
{\caption{Framework for combining deep learning (DL) with structural features (ORB) to provide a structural forecast via two pathways. Final forecasts of intensity (bottom-right) are based on both the observed intensity (top-right) and the observed and forecasted structure (center). The intermediate structural forecast is the key contrast with traditional end-to-end DL methods.}\label{fig:flowchart}}
{\includegraphics[width=.7\textwidth]{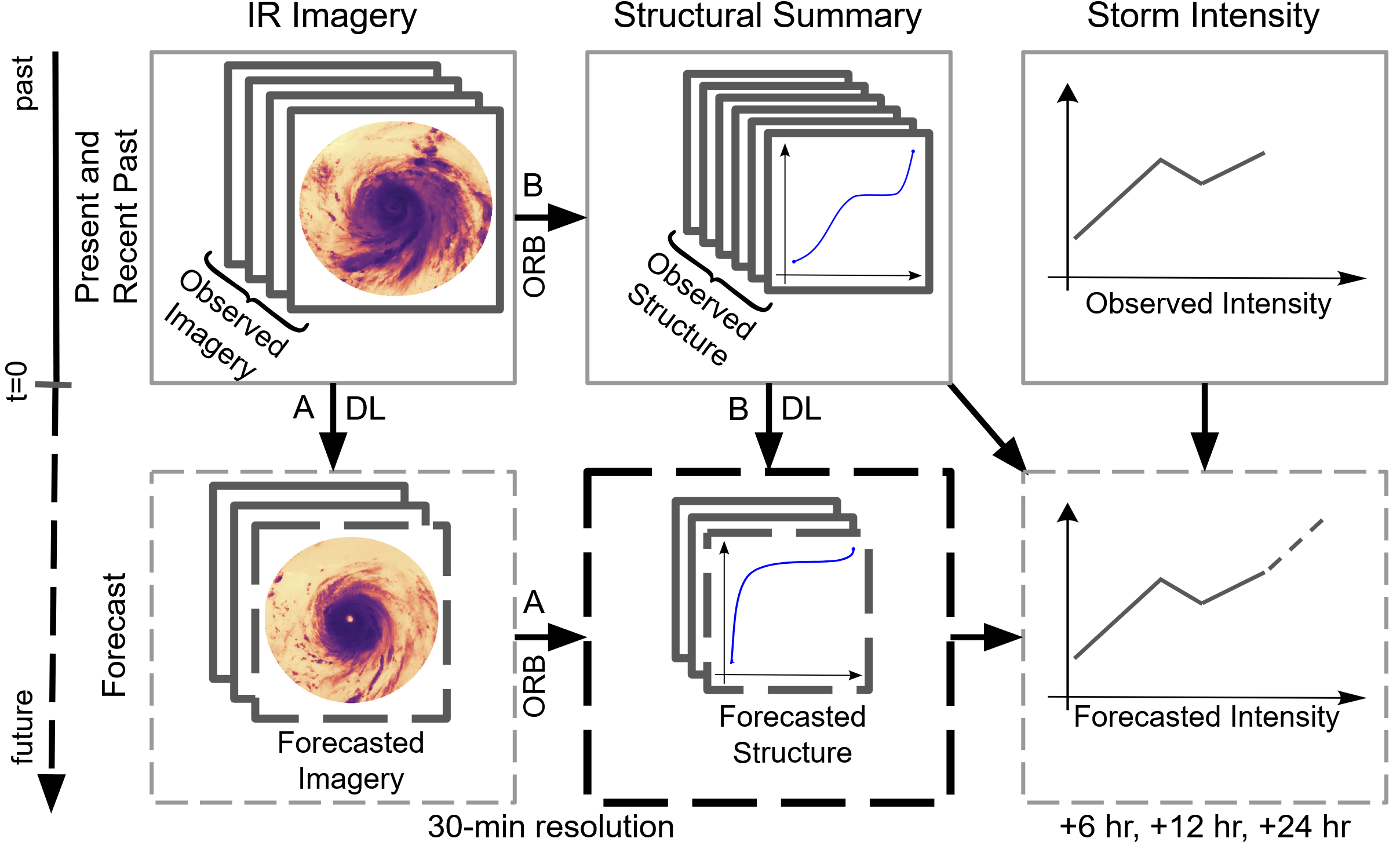}}
\end{figure}

Figure~\ref{fig:flowchart} describes our approach. In this figure, the top row indicates observed information, while the bottom row indicates forecasts; the final intensity forecast (bottom-right) depends on the past intensity (top-right) and a ``structural forecast'' (bottom-center) --- the predicted short-term evolution of TC structure itself. We obtain structural forecasts via two paths (A and B). Each approach extracts a set of interpretable features and uses DL to propagate the structural state of the TC into the near future. These parallel pathways allow forecasters to check the structural forecasts for agreement via quantitative error metrics in addition to checking them for physical plausibility directly examining both forecasted imagery and structure. Path A evolves the original IR imagery via DL and then computes structural features on the forecasted imagery. Path B instead computes structural features first then evolves them directly via DL. Traditional end-to-end DL approaches step directly from the observed IR imagery and intensity history to an intensity forecast; our framework's inclusion of interpretable structural forecasts as an intermediary step sets it apart from these traditional machine learning applications.

\section{Data} 
Geostationary satellite observations, including the Geostationary Operational Environmental Satellites (GOES), provide high spatial ($\le$4 km) and temporal ($\le$30 min) resolution imagery of the Atlantic and Pacific TC basins. Such observations are consistently available, unlike aircraft reconnaissance or land-based instruments such as radar. We will focus on longwave infrared ($\sim$10.7 $\mu$m), which provides estimates of cloud-top temperature. Because temperature generally decreases with height in the troposphere, low cloud-top temperatures typically indicate regions of stronger thunderstorms and thus deeper convection. The data collected consist of $\sim$200,000 IR images ($\sim$400x400 pixels) from 656 unique TCs in the North Atlantic and eastern North Pacific between 2000 and 2019. Storm location and intensity are drawn from the NHC's HURDAT2 database \cite{Landsea2013}.

\section{Providing Insight to TCs via Deep Learning and Forecasted Structure}
The ultimate goal of this work is to leverage evolving spatial structures in TCs to better understand storm behavior, specifically short-term storm intensity change (6- to 24-hour time frame). We would like to answer the question {\it ``Can we predict short-term intensity change using interpretable features from GOES-IR?"}

\textbf{Preliminary Work.} In prior work \cite{2020unlocking}, we have shown that a set of interpretable structural features drawn from GOES-IR imagery can nowcast (that is, predict ongoing) rapid intensity changes as well as SHIPS environmental predictors. Our approach extracted a set of ORB functions summarizing the Organization, Radial structure, and Bulk morphology of a TC image as continuous functions (of, e.g., the radius $r$ from the TC center or the threshold $c$  of level sets). The ORB functions were then compressed via a principal component analysis (PCA), and the PC coefficients served as inputs to logistic lasso models which classified the storms as rapidly changing in intensity or not. The (generalized) linear model allowed the user to directly relate the probability of a rapid TC intensity change to changes in key physical predictors like the eye-eyewall structure and symmetry of deep convection in the storm core. Our initial ORB suite performed as well as a subset of area-averaged SHIPS environmental predictors, while the combination of the two sets outperformed SHIPS alone. For reliable intensity forecasting, however, we need a richer suite of ORB functions, as well as a tool for projecting high-dimensional ORB functions $X_t \in \mathbb{R}^d$ (where $d$ is very large) into the future without a prior dimension reduction.

\textbf{Structural Summaries.} In the proposed work, we will develop a richer suite of structural summary functions that also includes center-independent structural features, measures of the spatial structure of vector fields, and additional satellite observations such as water vapor ($\sim6.5\mu$m) imagery. Crucially, we will compare our final intensity forecasts based on these structural features with intensity forecasts resulting from end-to-end DL models; in this way, we can assess the richness of the feature suite and quantify information lost in compression w.r.t. the root mean squared (RMS) intensity prediction errors. Instead of focusing on exhaustive feature design, we will predominantly explore new classes of features to improve the range of physical structures quantified by ORB. 

\textbf{Structural Forecasting.} In lieu of an end-to-end deep learning model, we will project our structural summaries of the TC 6, 12, and 24 hours into the future via DL; instead of only answering “How strong will the TC be in 6/12/24 hours?” we will also model “What will the TC look like in 6/12/24 hours?” This approach will provide the critical next step in the prediction pipeline, which enables forecasters to examine the structural forecast before relying on the intensity model. Since they can compare the forecast to the original satellite imagery, this improves stakeholder trust in the model, providing additional clues to the emergence of unrealistic TC structures or trajectories. Furthermore, this framework mimics the utilization of numerical weather models by statistical-dynamical intensity guidance; such models use physical laws rather than statistical learning to forecast the state of the atmosphere, but the final intensity guidance (e.g., SHIPS) still draws on summaries of this atmospheric forecast in the same way we will draw on a structural forecast.

We will generate structural forecasts via two pathways. The first of these (in Figure~\ref{fig:flowchart}, pathway A) steps the TC imagery forward in time, then computes ORB functions at each time step. This could be achieved using convolutional neural network architectures used in frame-to-frame prediction (as in \cite{babaizadeh2017variationalvideopred, franceschi2020residualvideopred}; see \cite{oprea2020videopredreview} for an exhaustive review). As video prediction tasks can be challenging, a modified version of this pathway could have deep convolutional architectures forecast intensity directly \cite{racah2017TCdetection, Kim2019hurricanetracker, monteleoni2020hurricanepred}; i.e., an end-to-end model going straight from the top left to the bottom right of Figure~\ref{fig:flowchart}. The second route (in Figure 1, pathway B) will directly step the ORB features forward in time, thus optimizing for the outcome of interest (TC structure). While linear auto-regressive models such as ARIMA \cite{box1976arima} are appropriate, we plan on leveraging recurrent neural networks \cite{Elman90findingstructure}, which have been shown to be successful in multivariate time series prediction tasks \cite{yu2017timeseriestensor, qin2017attentiontimeseries, lai2018lstnet, rangapuram2018deepstatespace} (see \cite{lim2020timesurvey} for a recent review). Comparing the output of pathways A and B will allow us to compare the impact of using interpretable structural features on the accuracy of structural forecasts against those automatically extracted by deep learning models.

\textbf{From Structure to Intensity.} 
Traditional linear models have historically been attractive to forecasters and scientists for reasons of interpretability and good performance in low sample size settings. However, linear models often struggle to capture the complex, time-varying processes which drive these storms. We here propose an approach that leverages DL to handle the high-dimensional structural forecasting problem, while the relationship between structure and intensity is handled by additive models (generalized additive models \cite{hastie1990generalized}, sparse additive models \cite{ravikumar2009sparse}, etc.). Additive models are attractive due to their combination of high-capacity, ease of interpretation, and straightforward visualizations. Importantly, these additive models have access to a structural forecast; the heavy lifting of high-dimensional time series prediction is handled by the deep learning models. In addition to this regression problem, comparison of trajectories via spectral clustering will be used to identify modes of evolution \cite{von2007tutorial}. Such identification of similar trajectories in historical TCs can provide forecasters and scientists alike with analogous TCs against which to compare the evolution of new storms.

\textbf{Evaluation Metrics.} We will have two quantitative metrics for the performance of this framework. First, we will compare the resultant RMS intensity errors and bias at 6, 12, and 24 hours to both end-to-end DL and to the official NHC forecasts for those times. Second, we will compare the $L_2$-distance between the resultant structural forecasts from the two pathways (to assess “sufficiency” of tracking only ORB features) as well as distance from the ground truth structural features. We acknowledge that a model based solely on satellite imagery with limited atmospheric and oceanographic data will not outperform state-of-the-art intensity prediction schemes on its own. However, a successful model which relates structural evolution to intensity change will be an invaluable addition to the toolbox utilized by forecasters and scientists analyzing TCs.

\bibliographystyle{plain}
\bibliography{main_arxiv}

\begin{thebibliography}{10}

\bibitem{babaizadeh2017variationalvideopred}
Mohammad {Babaeizadeh}, Chelsea {Finn}, Dumitru {Erhan}, Roy~H. {Campbell}, and
  Sergey {Levine}.
\newblock Stochastic variational video prediction.
\newblock {\em arXiv e-prints}, page arXiv:1710.11252, October 2017.

\bibitem{box1976arima}
George.E.P. Box and Gwilym~M. Jenkins.
\newblock {\em Time Series Analysis: Forecasting and Control}.
\newblock Holden-Day, 1976.

\bibitem{cangialosi2020}
John~P. Cangialosi, Eric Blake, Mark DeMaria, Andrew Penny, Andrew Latto,
  Edward Rappaport, and Vijay Tallapragada.
\newblock {Recent Progress in Tropical Cyclone Intensity Forecasting at the
  National Hurricane Center}.
\newblock {\em Weather and Forecasting}, 35(5):1913--1922, 08 2020.

\bibitem{DeMaria1999}
Mark DeMaria and John Kaplan.
\newblock An updated statistical hurricane intensity prediction scheme
  ({SHIPS}) for the {A}tlantic and {E}astern {N}orth {P}acific basins.
\newblock {\em Weather and Forecasting}, 14(3):326--337, 1999.

\bibitem{DeMaria2014}
Mark DeMaria, Charles~R. Sampson, John~A. Knaff, and Kate~D. Musgrave.
\newblock Is tropical cyclone intensity guidance improving?
\newblock {\em Bulletin of the American Meteorological Society},
  95(3):387--398, 2014.

\bibitem{Dvorak1975}
Vernon~F. Dvorak.
\newblock Tropical cyclone intensity analysis and forecasting from satellite
  imagery.
\newblock {\em Monthly Weather Review}, 103(5):420--430, 1975.

\bibitem{Elman90findingstructure}
Jeffrey~L. Elman.
\newblock Finding structure in time.
\newblock {\em {Cognitive Science}}, 14(2):179--211, 1990.

\bibitem{franceschi2020residualvideopred}
Jean-Yves {Franceschi}, Edouard {Delasalles}, Micka{\"e}l {Chen}, Sylvain
  {Lamprier}, and Patrick {Gallinari}.
\newblock Stochastic latent residual video prediction.
\newblock Proceedings of the International Conference on Machine Learning,
  2020.

\bibitem{monteleoni2020hurricanepred}
Sophie Giffard-Roisin, Mo~Yang, Guillaume Charpiat, Christina Kumler~Bonfanti,
  Bal\'azs K\'agl, and Claire Monteleoni.
\newblock Tropical cyclone track forecasting using fused deep learning from
  aligned reanalysis data.
\newblock {\em Frontiers in Big Data}, 3:1, 2020.

\bibitem{hastie1990generalized}
Trevor~J Hastie and Robert~J Tibshirani.
\newblock {\em Generalized additive models}, volume~43.
\newblock CRC press, 1990.

\bibitem{hu2020short}
Liang Hu, Elizabeth~A Ritchie, and J~Scott Tyo.
\newblock Short-term tropical cyclone intensity forecasting from satellite
  imagery based on the deviation angle variance technique.
\newblock {\em Weather and Forecasting}, 35(1):285--298, 2020.

\bibitem{Kaplan2010}
John Kaplan, Mark DeMaria, and John~A. Knaff.
\newblock A revised tropical cyclone rapid intensification index for the
  {A}tlantic and {E}astern {N}orth {P}acific basins.
\newblock {\em Weather and Forecasting}, 25(1):220--241, 2010.

\bibitem{kaplan2015rii}
John Kaplan, Christopher~M Rozoff, Mark DeMaria, Charles~R Sampson, James~P
  Kossin, Christopher~S Velden, Joseph~J Cione, Jason~P Dunion, John~A Knaff,
  Jun~A Zhang, et~al.
\newblock Evaluating environmental impacts on tropical cyclone rapid
  intensification predictability utilizing statistical models.
\newblock {\em Weather and Forecasting}, 30(5):1374--1396, 2015.

\bibitem{Kim2019hurricanetracker}
S.~{Kim}, H.~{Kim}, J.~{Lee}, S.~{Yoon}, S.~E. {Kahou}, K.~{Kashinath}, and
  M.~{Prabhat}.
\newblock Deep-hurricane-tracker: Tracking and forecasting extreme climate
  events.
\newblock In {\em 2019 IEEE Winter Conference on Applications of Computer
  Vision (WACV)}, pages 1761--1769, 2019.

\bibitem{Knapp2018}
K.~R. Knapp and S.~L. Wilkins.
\newblock Gridded satellite~({G}rid{S}at) {GOES} and {CONUS} data.
\newblock {\em Earth System Science Data}, 10(3):1417--1425, 2018.

\bibitem{lai2018lstnet}
Guokun Lai, Wei-Cheng Chang, Yiming Yang, and Hanxiao Liu.
\newblock Modeling long- and short-term temporal patterns with deep neural
  networks.
\newblock In {\em The 41st International ACM SIGIR Conference on Research \&
  Development in Information Retrieval}, SIGIR '18, page 95–104, New York,
  NY, USA, 2018. Association for Computing Machinery.

\bibitem{Landsea2013}
Christopher~W. Landsea and James~L. Franklin.
\newblock {A}tlantic hurricane database uncertainty and presentation of a new
  database format.
\newblock {\em Monthly Weather Review}, 141(10):3576--3592, 2013.

\bibitem{lim2020timesurvey}
Bryan Lim and Stefan Zohren.
\newblock Time series forecasting with deep learning: A survey.
\newblock {\em arXiv preprint arXiv:2004.13408}, 2020.

\bibitem{2020unlocking}
Trey McNeely, Ann~B. Lee, Kimberly~M. Wood, and Dorit Hammerling.
\newblock Unlocking goes: A statistical framework for quantifying the evolution
  of convective structure in tropical cyclones.
\newblock {\em arXiv preprint arXiv:1911.11089}, 2020.

\bibitem{olander2019advanced}
Timothy~L Olander and Christopher~S Velden.
\newblock The advanced dvorak technique (adt) for estimating tropical cyclone
  intensity: Update and new capabilities.
\newblock {\em Weather and Forecasting}, 34(4):905--922, 2019.

\bibitem{oprea2020videopredreview}
Sergiu {Oprea}, Pablo {Martinez-Gonzalez}, Alberto {Garcia-Garcia}, John
  {Alejandro Castro-Vargas}, Sergio {Orts-Escolano}, Jose {Garcia-Rodriguez},
  and Antonis {Argyros}.
\newblock A review on deep learning techniques for video prediction.
\newblock {\em arXiv e-prints}, page arXiv:2004.05214, April 2020.

\bibitem{qin2017attentiontimeseries}
Yao Qin, Dongjin Song, Haifeng Cheng, Wei Cheng, Guofei Jiang, and Garrison~W.
  Cottrell.
\newblock A dual-stage attention-based recurrent neural network for time series
  prediction.
\newblock In {\em Proceedings of the 26th International Joint Conference on
  Artificial Intelligence}, IJCAI'17, page 2627–2633. AAAI Press, 2017.

\bibitem{racah2017TCdetection}
Evan Racah, Christopher Beckham, Tegan Maharaj, Samira Ebrahimi~Kahou, Mr.
  Prabhat, and Chris Pal.
\newblock Extremeweather: A large-scale climate dataset for semi-supervised
  detection, localization, and understanding of extreme weather events.
\newblock In I.~Guyon, U.~V. Luxburg, S.~Bengio, H.~Wallach, R.~Fergus,
  S.~Vishwanathan, and R.~Garnett, editors, {\em Advances in Neural Information
  Processing Systems 30}, pages 3402--3413. Curran Associates, Inc., 2017.

\bibitem{rangapuram2018deepstatespace}
Syama~Sundar Rangapuram, Matthias~W Seeger, Jan Gasthaus, Lorenzo Stella,
  Yuyang Wang, and Tim Januschowski.
\newblock Deep state space models for time series forecasting.
\newblock In S.~Bengio, H.~Wallach, H.~Larochelle, K.~Grauman, N.~Cesa-Bianchi,
  and R.~Garnett, editors, {\em Advances in Neural Information Processing
  Systems 31}, pages 7785--7794. Curran Associates, Inc., 2018.

\bibitem{ravikumar2009sparse}
Pradeep Ravikumar, John Lafferty, Han Liu, and Larry Wasserman.
\newblock Sparse additive models.
\newblock {\em Journal of the Royal Statistical Society: Series B (Statistical
  Methodology)}, 71(5):1009--1030, 2009.

\bibitem{schmit2017closer}
Timothy~J Schmit, Paul Griffith, Mathew~M Gunshor, Jaime~M Daniels, Steven~J
  Goodman, and William~J Lebair.
\newblock A closer look at the {ABI} on the {GOES-R} series.
\newblock {\em Bulletin of the American Meteorological Society},
  98(4):681--698, 2017.

\bibitem{von2007tutorial}
Ulrike Von~Luxburg.
\newblock A tutorial on spectral clustering.
\newblock {\em Statistics and computing}, 17(4):395--416, 2007.

\bibitem{wood2015definition}
Kimberly~M Wood and Elizabeth~A Ritchie.
\newblock A definition for rapid weakening of {N}orth {A}tlantic and {E}astern
  {N}orth {P}acific tropical cyclones.
\newblock {\em Geophysical Research Letters}, 42(22):10--091, 2015.

\bibitem{yu2017timeseriestensor}
Rose {Yu}, Stephan {Zheng}, Anima {Anandkumar}, and Yisong {Yue}.
\newblock Long-term forecasting using higher order tensor {RNNs}.
\newblock {\em arXiv e-prints}, page arXiv:1711.00073, October 2017.

\end{thebibliography}

\end{document}